\documentclass[letterpaper, 10 pt, conference]{ieeeconf}  
\IEEEoverridecommandlockouts                              

\usepackage{graphics} 
\usepackage{graphicx}
\usepackage{caption}
\usepackage{subcaption}
\usepackage{epsfig} 
\usepackage{mathptmx} 
\usepackage{times} 
\usepackage{amsmath} 
\usepackage{amssymb}  
\usepackage[ruled,linesnumbered]{algorithm2e}
\usepackage{algpseudocode} 
\usepackage{booktabs} 
\usepackage{gensymb}
\usepackage{float}
\usepackage{algorithm2e}
\SetArgSty{textup}
\usepackage{hyperref}
\usepackage[autostyle=false, style=english]{csquotes}
\MakeOuterQuote{"}
\usepackage{adjustbox}
\usepackage{scalerel}
\usepackage{multirow}
\usepackage{epstopdf}
\usepackage{atbegshi}
\usepackage{fancyhdr}
\usepackage{lipsum} 

\usepackage{color}
\usepackage{hyperref} 
\hypersetup{
    colorlinks,
    linkcolor={red},
    urlcolor={blue},
    linktoc=all,
    citecolor={red}
}   

\usepackage{hyperref}
           
\usepackage[dvipsnames]{xcolor}
\usepackage{pgf}
\usepackage{tikz} 
\usetikzlibrary{positioning, shapes.geometric, shapes,arrows,automata}

\pgfdeclarelayer{background}
\pgfdeclarelayer{foreground}
\pgfsetlayers{background,main,foreground}   

\usetikzlibrary{arrows.meta}
\tikzset{>={Latex[width=1.8mm,length=1.5mm]}}

\title{\LARGE \bf
Adaptive Sampling-based Particle Filter \\ for Visual-inertial Gimbal in the Wild
}   
\author{Xueyang Kang$^{1}$, Ariel Herrera$^{2}$, Henry Lema$^{2}$, Esteban Valencia$^{2}$, Patrick Vandewalle$^{1}$ 
\thanks{*This work was supported by VLIR-UOS project under agreement No. EC2020SIN278A101.} 
\thanks{$^{1}$ The authors are with PSI Department of Electrical Engineering (ESAT), KU Leuven, Belgium. Email: {\tt\small alex.kang@kuleuven.be}}%
\thanks{$^{2}$ The authors are with Department of Mechanical Engineering, Escuela Politecnica Nacional (EPN), Ecuador{\tt\small}}%
}
\def\xk{\textcolor{black}}

\begin{document}
\maketitle
\thispagestyle{empty}
\pagestyle{empty}
\section*{Abstract}
In this paper, we present a Computer Vision (CV) based tracking and fusion algorithm, dedicated to a 3D printed gimbal system on drones flying in nature. The whole gimbal system can stabilize the camera orientation robustly in challenging environments by using skyline and ground plane as references. Our main contributions are the following: a) a light-weight Resnet-18 backbone network model was trained from scratch, and deployed onto the Jetson Nano platform to segment the image specifically into binary parts (ground and sky); b) our geometry assumption from the skyline and ground cues delivers the potential for robust visual tracking in the wild by using the skyline and ground plane as references; c) a manifold surface-based adaptive particle sampling can fuse orientation from multiple sensor sources flexibly. Jetson Nano tests the whole algorithm pipeline on our 3D-printed gimbal module. The experiments were performed on top of a building in a real landscape. The public code link: \href{https://github.com/alexandor91/gimbal-fusion.git}{https://github.com/alexandor91/gimbal-fusion.git}.
\section{Introduction}
Gimbal platforms have been widely used in photogrammetry and robot perceptual modules to stabilize the camera pose, thereby improving the captured video quality. Usually, a gimbal is mainly composed of sensors and actuators. The sensor's orientation measurements can be utilized directly by an actuator to steer the camera toward the desired pose. However, off-the-shelf custom gimbals are either quite expensive or depend on a high-precision IMU or a brushless DC motor with a Hall sensor to read angles, which is prone to noise drift over long-term operation. The goal of our project is to deploy a gimbal platform carried by a UAV for volcanic eruption surveillance in an unpopulated region full of mountains. Hence, our work differs from prior work in that a robust platform should be devised, in operation over a long term and with endurance in the challenging wild. \xk{The main contribution of this paper is threefold.
\begin{itemize}
    \item A lightweight binary segmentation model is trained to label the ground and sky pixels specifically, aiming for real-time inference on the embedded device. 
    \item Natural cues in the challenging mountainous region, such as the ground plane and skyline isolated from the aforementioned binary mask, are utilized as references for the gimbal stabilization to infer the rotation angles.
    \item To fuse the roll and pitch rotation angles from multiple sensor modalities, i.e., IMU and CV pipeline, a non-linear particle filter at varying resolutions over the manifold surface is proposed, and it is implemented on the Jetson Nano board with a demo test. 
\end{itemize}
} 
\section{Related work}
The problem of video stabilization can be traced back to the Electronic Image Stabilization (EIS) technique \cite{EIS}, which relied on handcrafted feature points in consecutive image frames to search for correspondences for image alignment. Feature point-based tracking \cite{surf-tracking}, or optical flow tracking \cite{optical-flow}, required distinctive feature points or consistent raw pixel intensity distributions across video frames as shown in the work by Kulkarni et al. \cite{video-stabilise}. Thus, it is sensitive to pixel noise induced by motion blur. On the other hand, the humanoid robot vision system \cite{robot IMU}, or complex video task \cite{hybrid-motion} compensated for the rotation and shift of video images by using orientation from IMU directly. A good review of traditional motion prediction work applied to video stabilization can be found in a study by Rawat et al. \cite{review-stabilization-tech}.
\begin{figure}[H]
\centering
\includegraphics[width=0.48\textwidth]{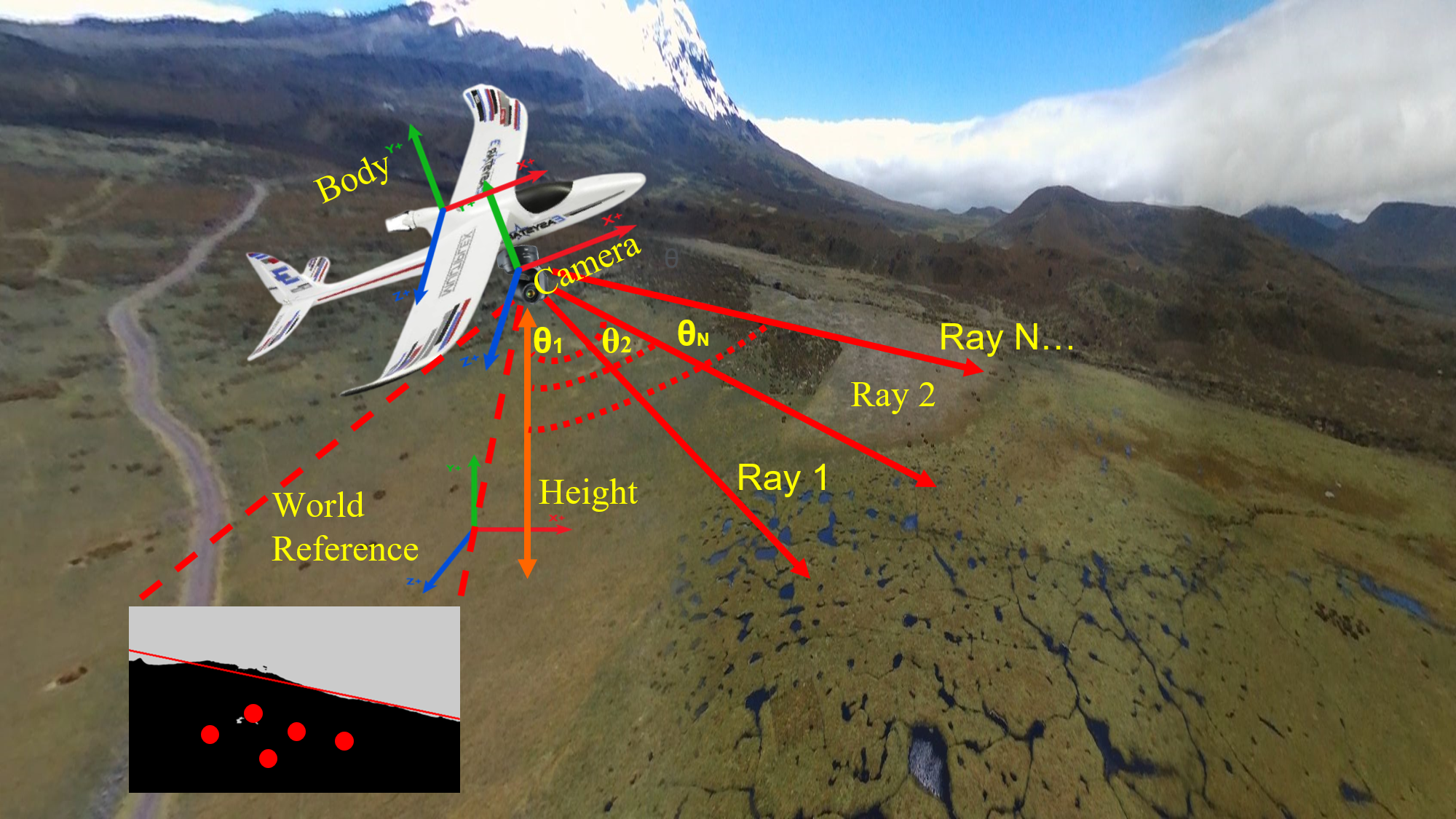}
\caption{Demo of gimbal platform on the fixed-wing airplane.}
\label{fig:overview}
\vspace{-1.0em}
\end{figure}
Video stabilization can also be considered as an image deblurring task. The main goal is to infer the pixel intensity in a blurry area from neighboring pixels in single or multiple images, incorporating both temporal and spatial constraints from the video. Recent deep learning techniques boost video processing performance by introducing a prior from a dataset in specific domains, e.g.,  Jiyang Yu et al. \cite{selfie-tracking} warped the selfie image patch locally by decoding the scene representation from multiple encoder outputs, such as foreground contours, feature points on the background, and 3D facial mesh. Chen Li et al. \cite{deep IMU stabilization} trained a motion estimation model by using raw IMU data to predict the visual odometry in a sliding time window, similar to a filter approach. 

The video stabilization was integrated into a large mapping and localization framework via graphs in a self-supervised manner by Lee et al. \cite{stabilization-depth}. To tackle the problem of video stabilization in dynamic scenes, a dense warping field as scene representation was trained from consecutive video frames by Liu et al. \cite{hybrid-neural-fusion}, then the warped parts are blended to synthesize the stabilized image. The framework by J. Choi et al. \cite{self-supervised} even made use of a motion prediction model based on optical flow tracking. However, all those models involve a lot of processing overhead and are quite heavy to deploy on an edge device in real time. 
Video stabilization was also widely applied in aerial surveillance by \cite{video-surveillance, Patent, Patent1}, or UAV attitude stabilization as the work done by Chung-Cheng Chiu et al. \cite{v-small-uav}. Using natural cues to estimate the orientation of UAVs or fixed-wing airplanes was performed in many real applications \cite{sky},\cite{hd},\cite{fixed-wing}, all relying on skyline tracking. However, they require delicate tuning to search for a boundary between sky and ground, even input images captured under ideal conditions. Other works depend on the tracking of geometric primitives on 2D images or in 3D, either by discrete feature points \cite{sift-video}\cite{video-stablize2}\cite{surf-tracking}, or a curved boundary \cite{boundary}. Five-point algorithm \cite{five-point} or curvature alignment can be used to predict the ego-pose of the camera from feature correspondences. Moving objects were detected in the image view through a tracking filter proposed by Ahlem Walha et al. \cite{uav-tracking}, over the "SIFT" feature points \cite{sift}. However, in a natural setting, there are a lot of spurious features, such that tracking algorithms relying on feature points may fail.
\section{Overview}
 As shown in Figure \ref{fig:overview}, the gimbal system is placed underneath an airplane body including a camera, IMU, and barometer. To overcome the aforementioned correspondence issues of feature points, the gimbal system can take advantage of natural cues such as skyline and ground to stabilize the camera pose. The bottom of Figure \ref{fig:overview} exhibits a binary mask with a skyline and points in the ground region. Rays passing through the red dots along with the height from the barometer, allow us to determine the 3D position of the ground plane through trigonometry. The hardware and software diagrams are presented.
\subsection{Open source Hardware}
The gimbal platform design is based on open-source hardware. The main processing unit is a Jetson Nano, featured with 2GB GPU memory and Quad-core ARM A57, connected to IMU, camera, and barometer sensors. An OpenCR 2.0 driver board maps the driving command to control commands for two servo motors.
\begin{figure}[H]
\centering
\includegraphics[width=0.43\textwidth]{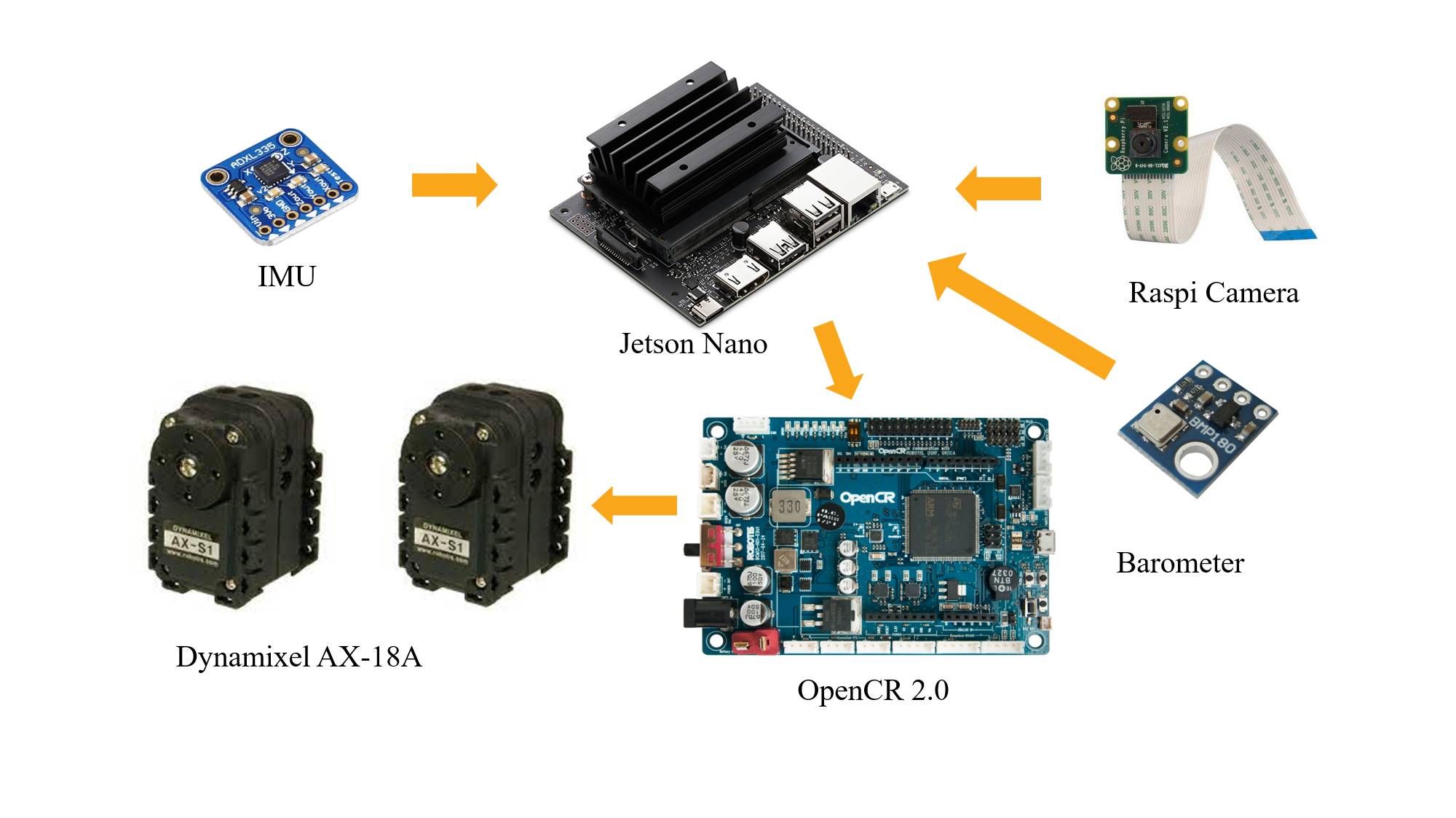}
\label{fig:hardware}
\vspace{-2.0em}
\caption{Open source hardware setup.}
\vspace{-1.0em}
\end{figure}

\subsection{ROS nodes}
The software is composed mainly of three parts: the preprocessing part including the network model and geometric primitive extraction, followed by a tracking module to align the skyline and normal of the ground plane in the current frame with those in the reference frame. The compensation angles from various pipelines are then fed into the proposed particle filter presented in Section \ref{section-pf} to obtain fusion orientations, further as input for the controller to stabilize the camera. 

\begin{figure}[H]
  \centering
  \resizebox{.48\textwidth}{!}{
  \begin{tikzpicture}
    \node[draw,
        circle,
        text width=1.0cm,
        minimum size=0.1cm,
        fill=Rhodamine!40,
        text centered] (Video){\centering Video\\Input};
    
    \node [draw,
        fill=Goldenrod,
        minimum width=1cm,
        minimum height=1cm,
        above=0.5cm of Video,
        text centered
    ]  (SegNet) {\begin{tabular}{c} Binary Segmentation \\ \& Skyline Isolation \end{tabular}};
    
    \node[draw,
        circle,
        text width=1.0cm,
        minimum size=0.4cm,
        fill=Rhodamine!40,
        right=0.2cm of SegNet,
        text centered
    ] (Skyline){Skyline};

    \node[draw,
        text centered,
        circle,
        text width=1.0cm,
        minimum size=0.2cm,
        fill=Rhodamine!40,
        above=0.6cm of Skyline
    ] (Ground){\centering Ground\\Plane};

    \node[draw,
        circle,
        text width=0.9cm,
        minimum size=0.1cm,
        fill=Rhodamine!40,
        right=0.5cm of Video,
        text centered
    ] (IMU){\centering IMU\\Data};

    \node [draw,
        fill=Goldenrod,
        minimum width=1cm,
        minimum height=1cm,
        right=0.2cm of Skyline,
        text centered
    ]  (Perception) {\begin{tabular}{c} Geometry\\Tracking \end{tabular}};

    \node[draw,
        circle,
        text width=0.9cm,
        minimum size=0.1cm,
        fill=Rhodamine!40,
        above=0.8cm of Perception,
        text centered
    ] (Barometer){Height};

    \node [draw,
        fill=Goldenrod,
        minimum width=0.8cm,
        minimum height=0.8cm,
        below=0.75cm of Perception,
        text centered
    ]  (pf) {\begin{tabular}{c}Particle\\Filter \end{tabular}};
    
    \node[draw,
        circle,
        text width=0.9cm,
        minimum size=0.1cm,
        fill=Rhodamine!40,
        right=0.8cm of pf,
        text centered
    ] (SO2){\centering Angles\\SO(2)};
    
    \node [draw,
        fill=Goldenrod,
        minimum width=1cm,
        minimum height=1cm,
        above=0.6cm of SO2,
        text centered
    ]  (Controller){\begin{tabular}{c}Gimbal\\Controller\end{tabular}};
    
    \node[draw,
        circle,
        text width=0.9cm,
        minimum size=0.1cm,
        fill=Rhodamine!40,
        above=0.5cm of Controller,
        text centered
    ] (Motor){\centering Servo\\Motor};
              
    \draw[->] (Video.north) -- (SegNet.south);
    \draw[->] (SegNet.east) -- (Skyline.west);
    \draw[->] (SegNet.north) -- (Ground.west);
    \draw[->] (Skyline.east) -- (Perception.west);  
    \draw[->] (Ground.east) -- (Perception.north);  
    \draw[->] (IMU.east) -- (pf.west);   
    \draw[->] (Perception.south) -- (pf.north); 
    \draw[->] (Barometer.south) -- (Perception.north);
    \draw[->] (pf.east) -- (SO2.west);   
    \draw[->] (SO2.north) -- (Controller.south);   
    \draw[->] (Controller.north) -- (Motor.south);

     
     
     
  \begin{pgfonlayer}{background}
    \path (SegNet.west |- SegNet.north)+(-0.2,0.4) node (A) {};
    \path (Perception.east |- Perception.south)+(+0.1,-0.4) node (B) {};
    \path[fill=yellow!20,rounded corners, draw=black!50, dashed] (A) rectangle (B);
  \end{pgfonlayer}
  \end{tikzpicture}}
  \caption{Block diagram of the presented algorithm. Circular nodes in pink are signals or controlled targets. Rectangular boxes in yellow are ROS nodes for the algorithm, the dashed region is the front-end perception part, including tracking of skyline and ground plane.}
  \label{fig:ros-nodes}
\vspace{-0.7em}
\end{figure}
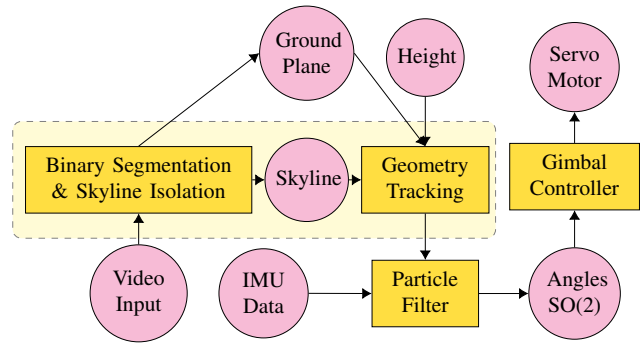  
\section{Perception}
\label{section:percep}
The perception part is structured into preprocessing and rotation estimation, where the rotation estimation can be further separated into two pipelines: roll and pitch prediction from skyline tracking and rotation estimation from ground plane tracking. The following subsections follow this structure.

\subsection{Preprocessing}
\begin{figure}[H]
\centering
\begin{subfigure}{.48\textwidth}
  \centering
  \includegraphics[width=1.0\textwidth]{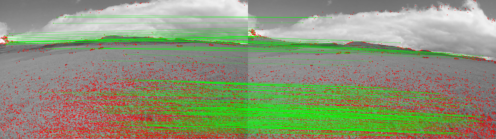}
    \caption{\centering Correspondences based on "SIFT" \cite{sift} feature points of neighboring frames.}
    \label{fig:feature-association}
\end{subfigure}
\medskip
\begin{subfigure}{.24\textwidth}
  \centering
  \includegraphics[width=1.0\textwidth]{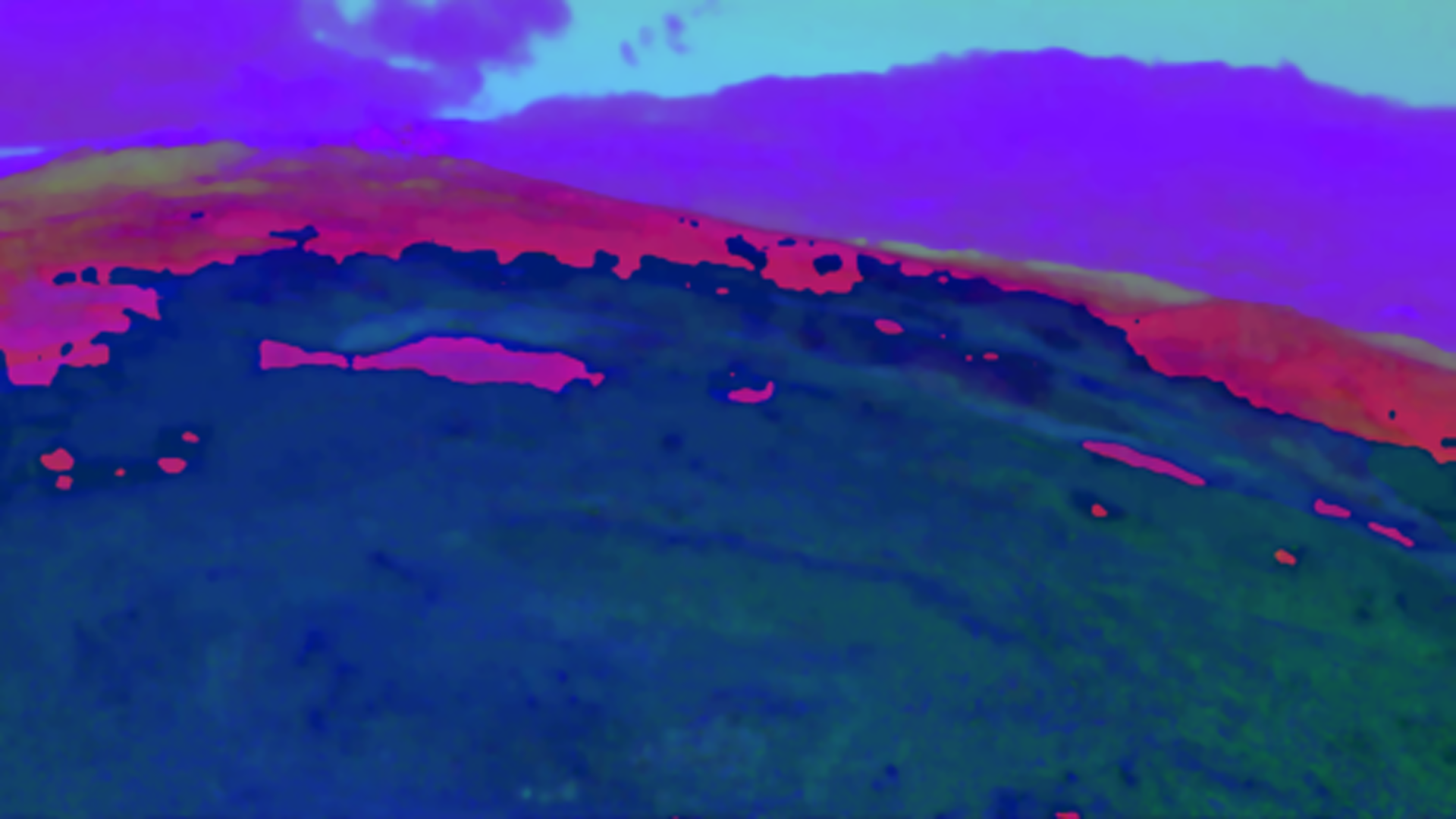}
    \caption{\centering HSV image converted from a raw RGB image.}
    \label{fig:hsv-img}
\end{subfigure}\hfil
\begin{subfigure}{.24\textwidth}
  \centering
  \includegraphics[width=1.0\textwidth]{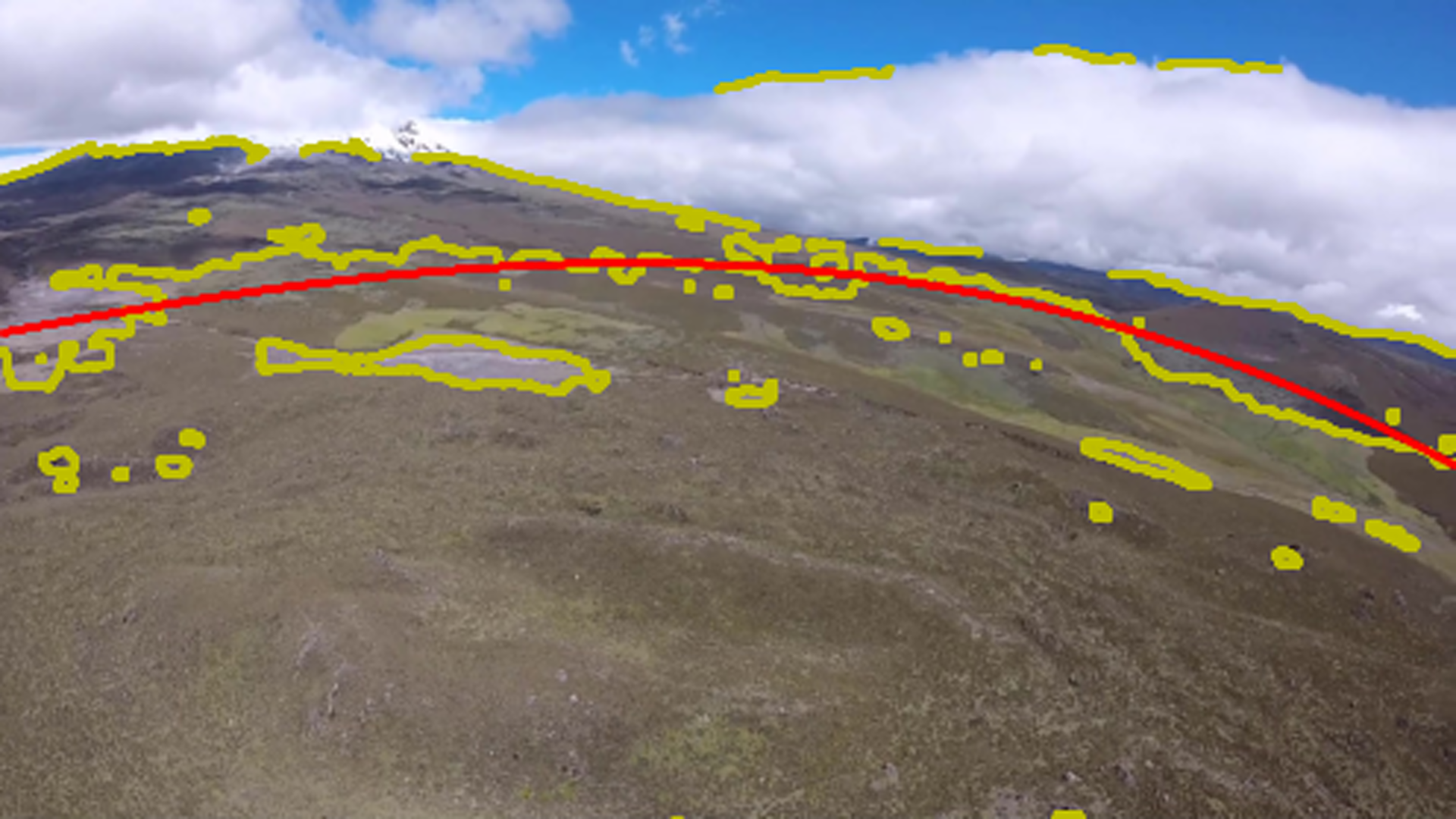}
    \caption{\centering Boundary curve-fitting on the detected Canny edges.}
    \label{fig:curve-fitting}
\end{subfigure}
\vspace{-1.0em}
    \caption{Failure case demo using OpenCV pipeline.}
    \label{fig:failure-case}
\vspace{-1.5em}
\end{figure}

\begin{figure}[H]
\begin{subfigure}{.15\textwidth}
  \centering
  \includegraphics[width=1.0\linewidth]{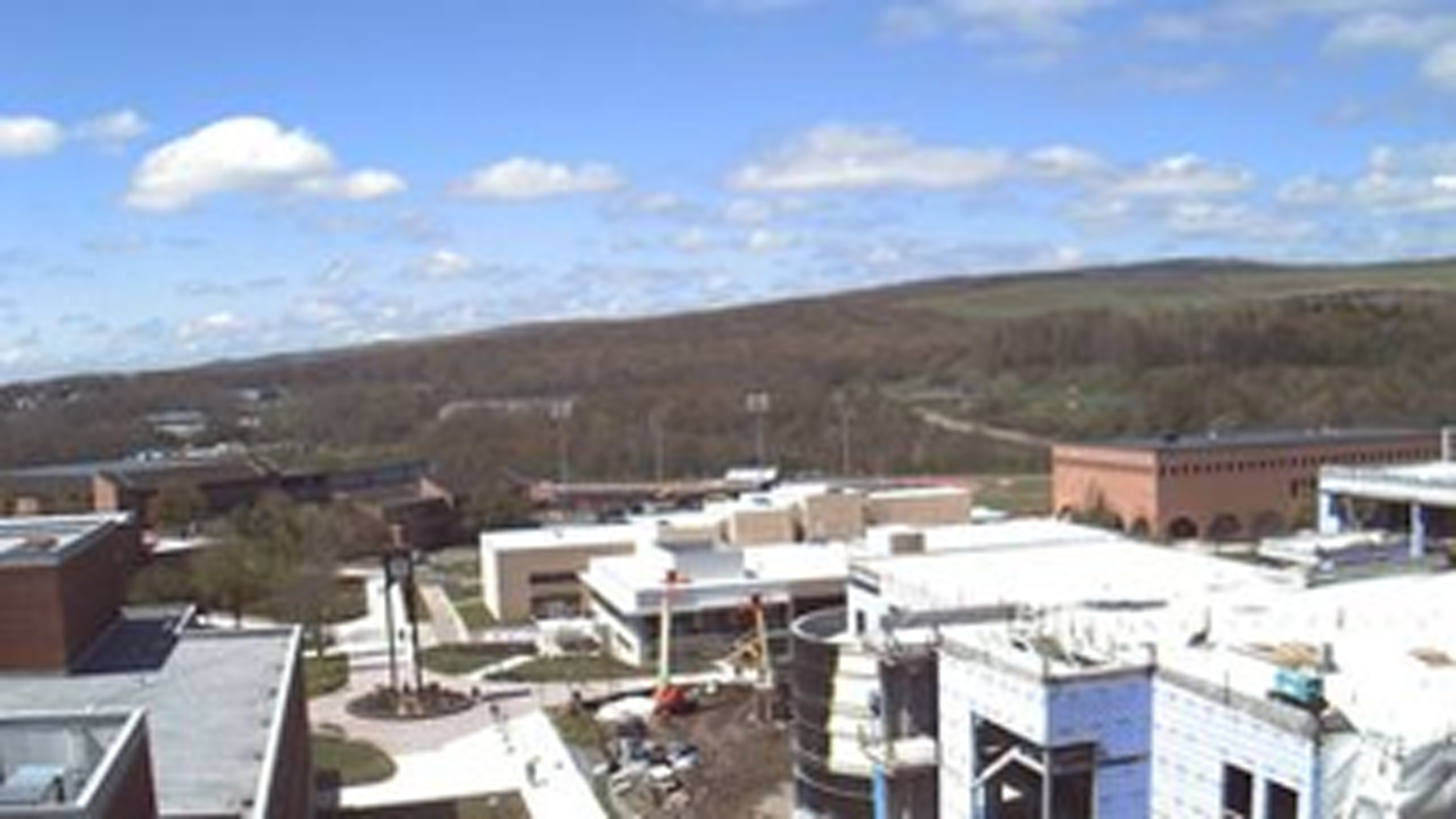}
\end{subfigure}\hfil
\begin{subfigure}{.15\textwidth}
  \centering
  \includegraphics[width=1.0\linewidth]{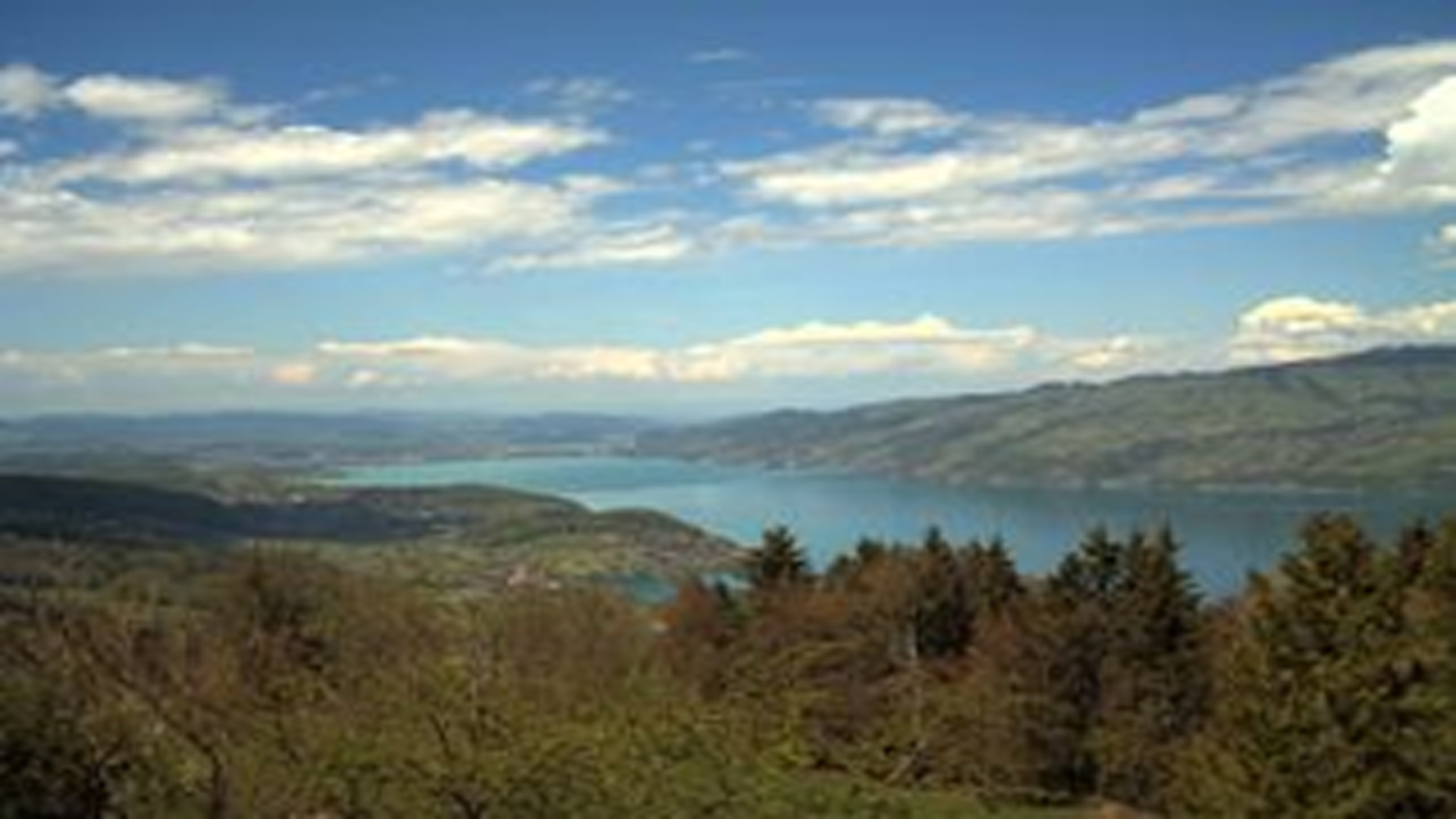}
\end{subfigure}\hfil
\begin{subfigure}{.15\textwidth}
  \centering
  \includegraphics[width=1.0\linewidth]{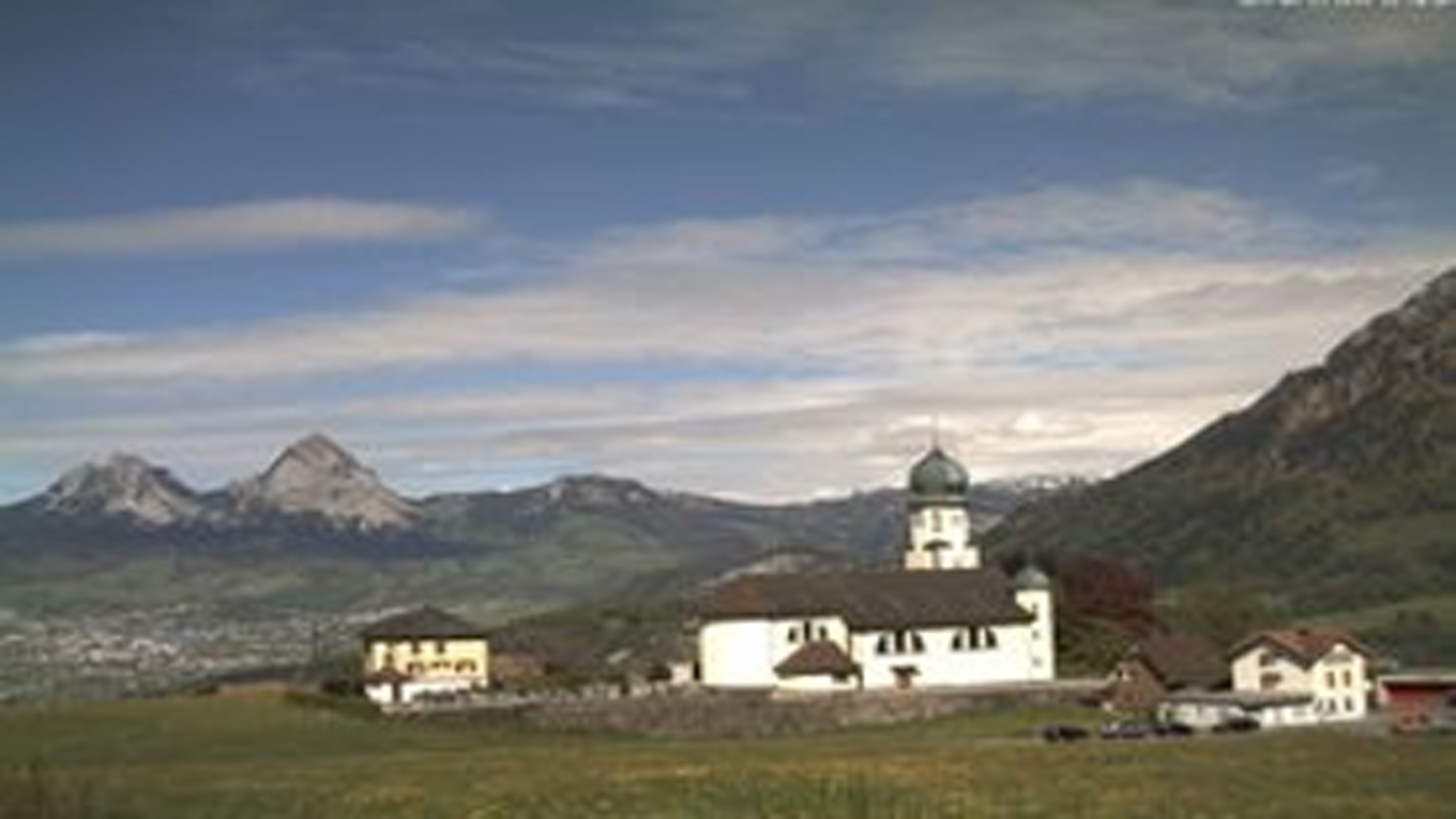}
\end{subfigure}
\medskip

\begin{subfigure}{.15\textwidth}
  \centering
  \includegraphics[width=1.0\linewidth]{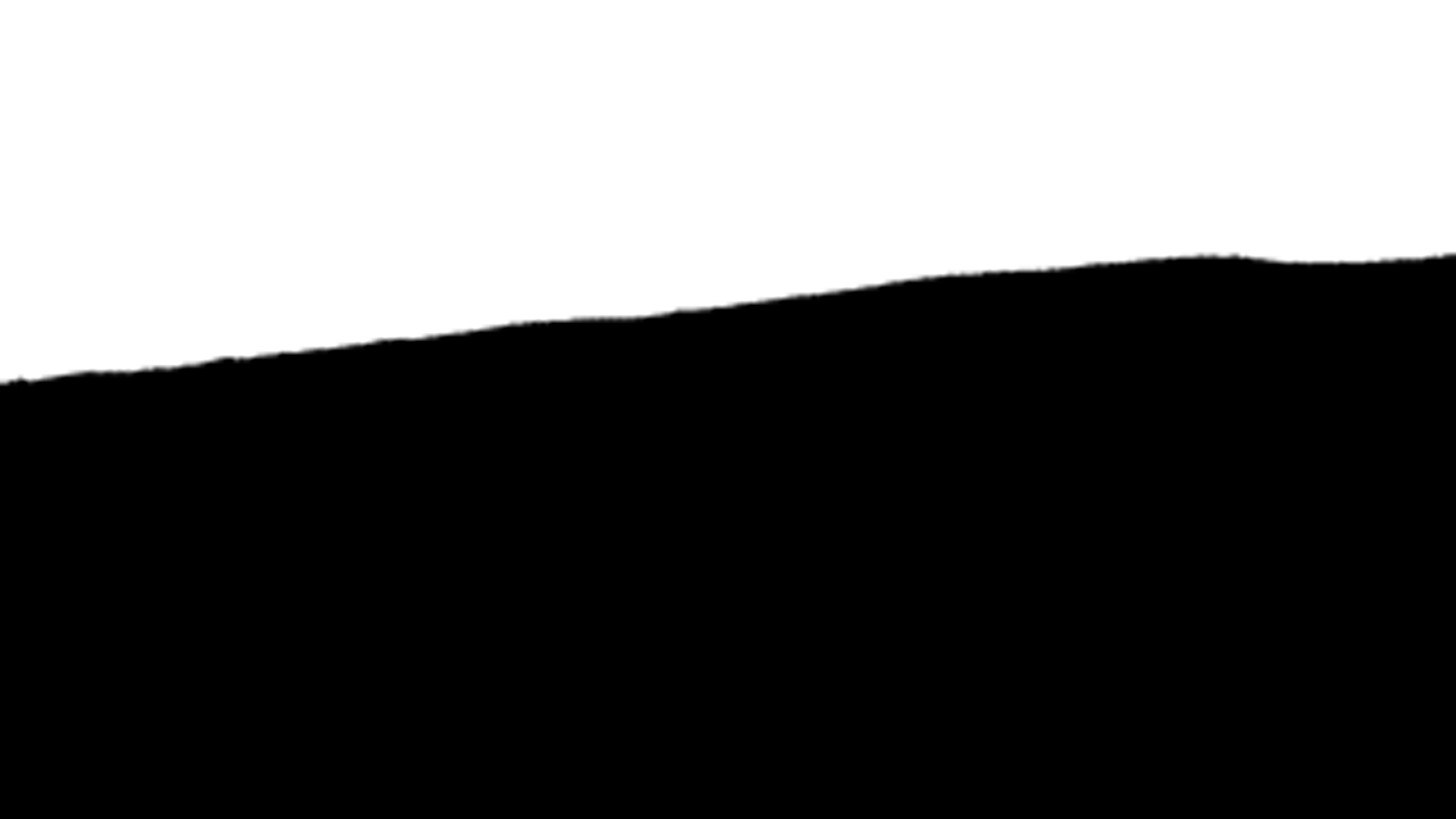}
\end{subfigure}\hfil
\begin{subfigure}{.15\textwidth}
  \centering
  \includegraphics[width=1.0\linewidth]{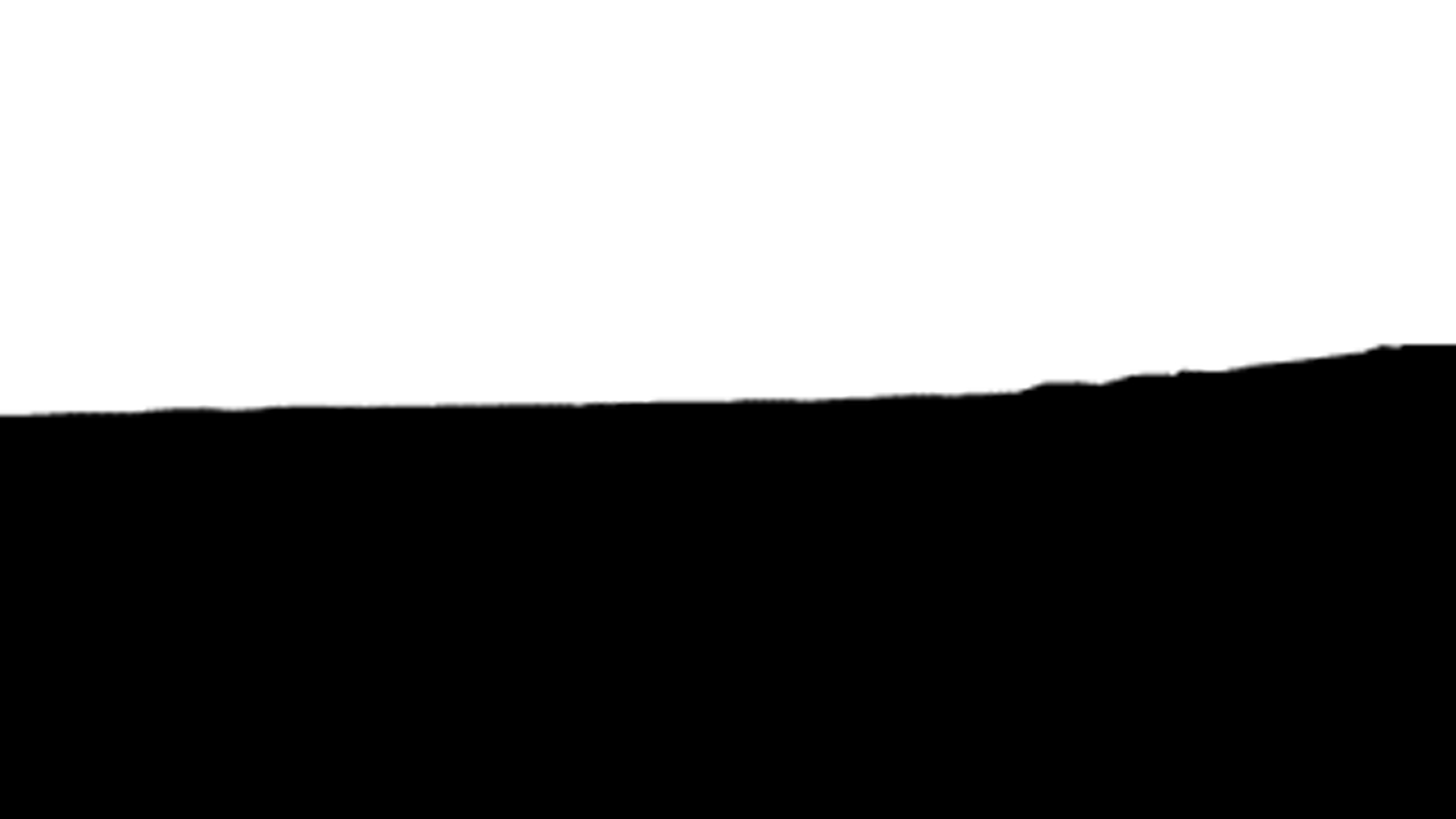}
\end{subfigure}\hfil
\begin{subfigure}{.15\textwidth}
  \centering
  \includegraphics[width=1.0\linewidth]{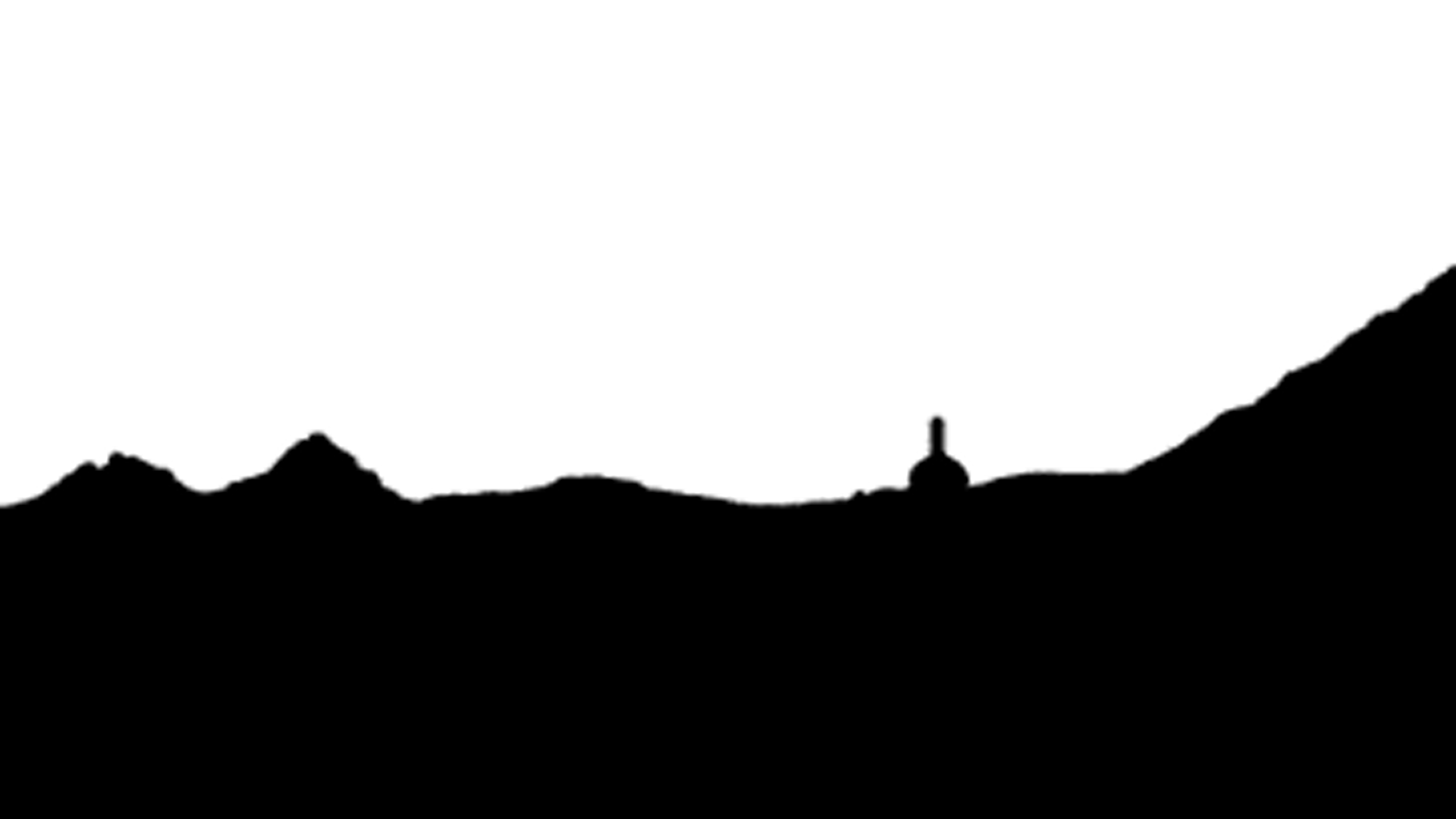}
\end{subfigure}
\caption{Sample images and ground truth masks for training.}
\label{fig:train-sample}
\vspace{-0.5em}
\end{figure}

\xk{We first tried the general computer vision processing pipelines, but they all failed due to spurious feature candidates. As illustrated in Figure \ref{fig:failure-case}, the similar appearances in the grassland region and cloudy sky all pose a great challenge to correspondence search. In Figure \ref{fig:feature-association}, many false correspondences are found. Canny edge detection \cite{canny} is applied to find the boundary between the ground and sky region on HSV image, nevertheless, some brightness of the sky is cast onto the grass ground, generating the wrong boundary in Figure \ref{fig:curve-fitting}. To tackle this challenging segmentation task, we finally choose the data-driven approach, by training a Resnet-18 \cite{resnet} network on "Skyfinder" dataset \cite{dataset} first, followed by a fine-tuning on the self-collected dataset with one hundred images. Binary cross entropy loss is applied for pre-training through 100 epochs, and fine-tuning with 30 epochs respectively. The total training time is less than two hours. Some training data samples along with ground truth masks are presented in Figure \ref{fig:train-sample}. The model is exported into "ONNX" and optimized by "TensorRT" to convert to "FP16" precision for Jetson Nano deployment. We found the model can achieve above 90\% success rate for the segmentation on average, only when under some extreme cases like overexposure, the failure may happen. Additionally, our use case is for mountainous terrain, the scenario with mirror effects and reflections by water on the ground is not advisable, due to the similar color distributions of the sky and the ground.}

\subsection{Skyline Tracking}
\begin{figure}[H]
\begin{subfigure}{.24\textwidth}
  \centering
  \includegraphics[width=1.0\linewidth]{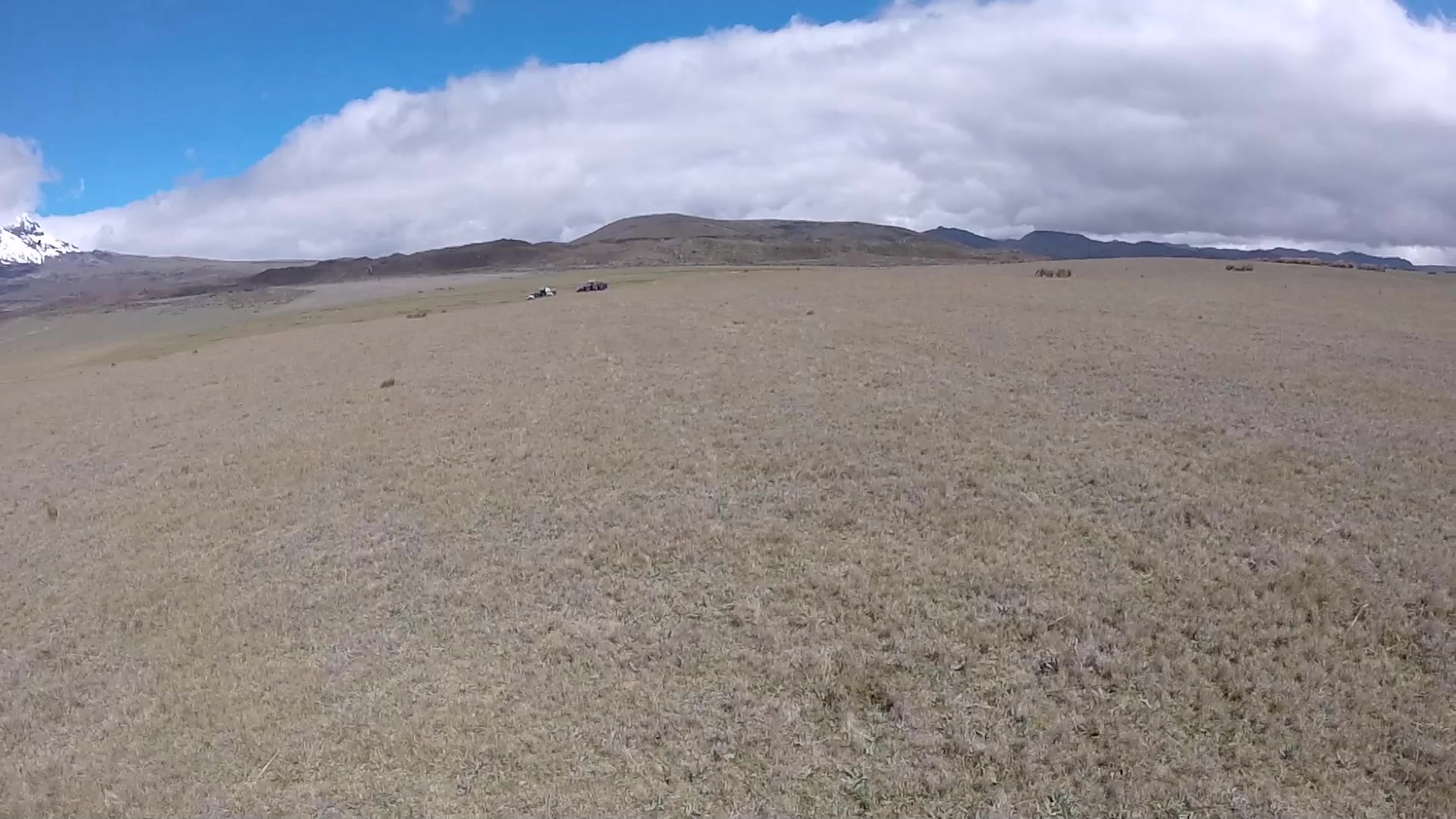}
  \caption{Reference image}
  \label{fig:reference-img}  
\end{subfigure}\hfil
\begin{subfigure}{.24\textwidth}
  \centering
  \includegraphics[width=1.0\linewidth]{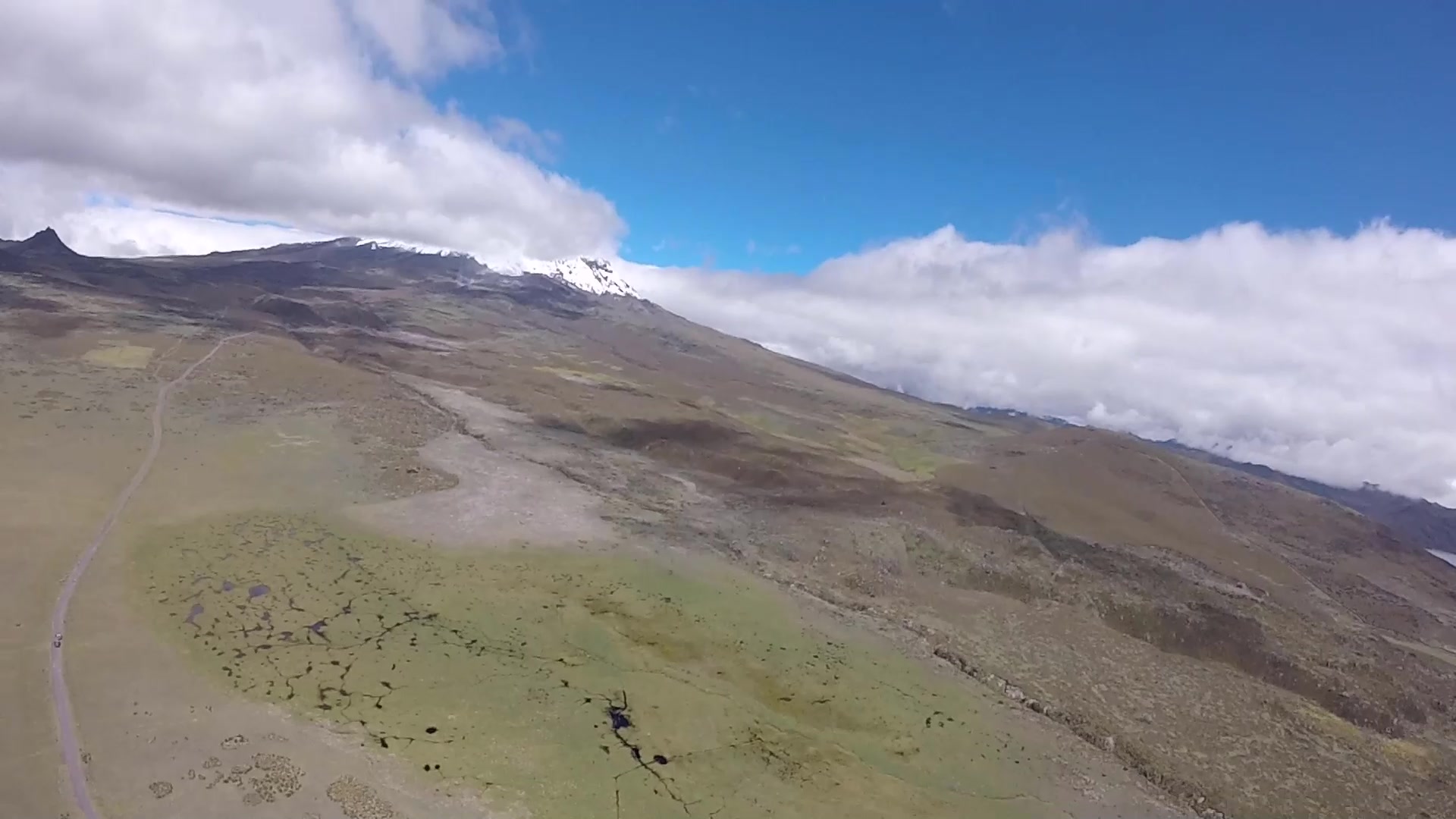}
  \small{\caption{Current image frame}}
  \label{fig:current-img}  
\end{subfigure}
\medskip

\begin{subfigure}{.24\textwidth}
  \centering
  \includegraphics[width=1.0\linewidth]{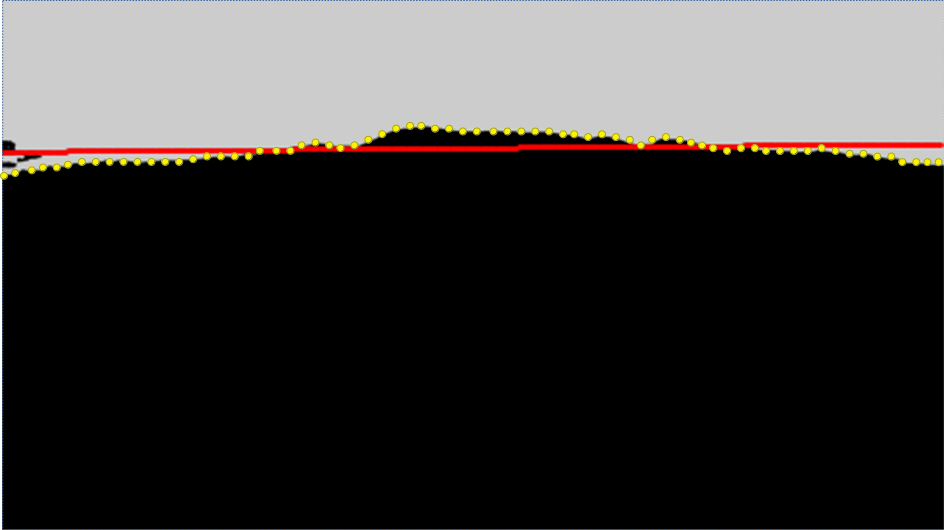}
  \caption{Reference mask with skyline}
  \label{fig:ref-mask}  
\end{subfigure}\hfil
\begin{subfigure}{.24\textwidth}
  \centering
  \includegraphics[width=1.0\linewidth]{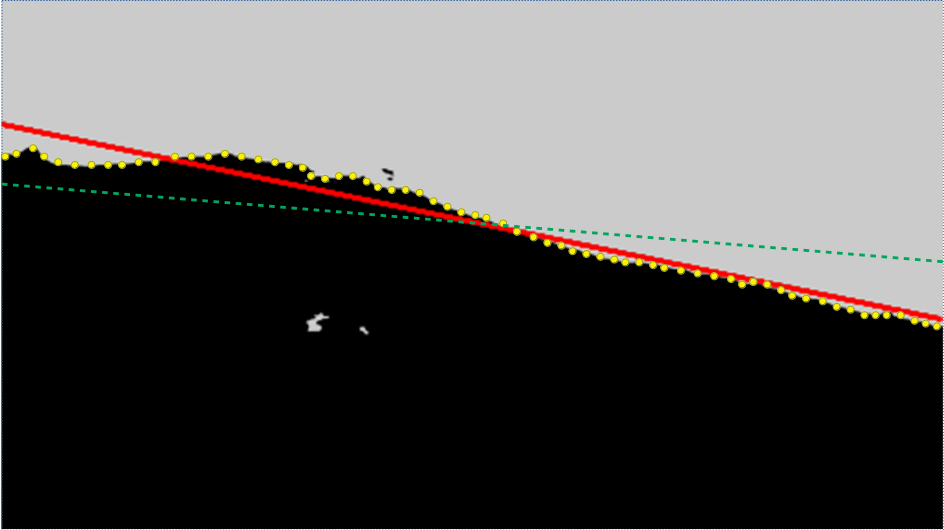}
  \caption{Current mask with skyline}
  \label{fig:curr-mask}  
\end{subfigure}
\caption{Segmented images for skyline search.}
\label{fig:seg-samples}
\vspace{-1.4em}
\end{figure}

\xk{Starting from the binary mask results gained from the network model, the skyline can be extracted along the boundary direction. The extracted skyline can be further considered as a cue to estimate the roll and pitch of the camera, as shown in Figure \ref{fig:seg-samples}. The boundary points, represented by yellow dots in Figure \ref{fig:ref-mask}, can be implemented firstly over the whole reference image. A straight line is fitted as illustrated in red (Figure \ref{fig:ref-mask}), using two parameters, slope $m'$ and intercept $b'$ in Equation 1. For the subsequent images, we use a constant angular velocity model derived from skyline tracking of the previous two frames to predict the skyline position in the current frame, like the green line presented in Figure \ref{fig:curr-mask}. The assumption is that the camera remains not upside down, due to the airplane maneuverability constraints. The boundary can always be searched along the vertical direction of the predicted skyline (green). In practice, the skyline points are further down-sampled to speed up the search. Once the current sample points of skyline are attained, the estimated skyline (red) in the current frame can be derived via least-square in Figure \ref{fig:curr-mask}. }
\begin{flalign}
m&'x + b' = y'\\
m&x+ b = y \\
\alpha &= \arctan(m) -  \arctan(m')\\
\beta &= \arctan\left(\frac{h^{}_{1}-c^{}_{y}}{f^{}_{y}}\right)-\arctan\left(\frac{h^{}_{2}-c^{}_{y}}{f^{}_{y}}\right)
\end{flalign}
\xk{The roll $\alpha$ and pitch $\beta$ angles can be derived from Equations 3, and 4 respectively by subtracting the angles. $c_{y}$ is half of the image height, and $h^{}_{1}$ and $h^{}_{2}$ are the height of the center point of skyline in the current image frame and reference frame respectively. $f^{}_{y}$ is the focal length $y$ of the camera.}
Roll and pitch have a specific tolerance range to avoid unnecessary operations, so processing is only triggered when the movement is out of this range. Roll angle can be predicted from the slope $m$ of the skyline, followed by pitch estimation, which is on top of the image result after roll compensation. There is a total of three cases in our system, that is pure roll, pure pitch, or both happening simultaneously. Here the height shift resulting from translation is subtracted before rotation processing by barometer readings.
\subsection{Estimation from Ground Plane Tracking}
Ground plane tracking relies on the normal vector of the ground plane, as demonstrated in Figure \ref{fig:overview}. A set of points in the ground region of the binary mask are sampled evenly, followed by a back projection to the camera frame corresponding to Equations 5-11.
\begin{flalign}
\mathbf{K} &= 
\begin{bmatrix}
f_x & 0 & c_x \\
0 & f_y & c_y \\
0 & 0 & 1
\end{bmatrix} \\
\mathbf{\rho_{i}} &= [u, v, 1]^{T} \\
\mathbf{P_{i}} &= \mathbf{{K}^{-1}} \mathbf{\rho_{i}}, i \in (1...N) \\
\mathbf{N^{}_{G}} &= [0, 0, g^{}_{z}]^{T}\\
\cos\theta &= \frac{\mathbf{P^{}_{i}} \cdot \mathbf{N^{}_{G}}}{||\mathbf{P^{}_{i}}||\cdot||\mathbf{N^{}_{G}}||} \\
l_{i} &= \frac{h}{\cos \theta} \\
\mathbf{P'_{i}} &= l_{i}\mathbf{P_{i}}
\end{flalign}
The height $h$ is measured from the barometer, and the ray direction passing through the pixel position is $\mathbf{P_{i}}$ on the left side of Equation 7. $\mathbf{K}$ is the intrinsic matrix obtained from calibration \cite{zhang}. $\theta$ is derived from the dot product of the gravitational vector and the ray direction. Length scale $l_i$ can be calculated by trigonometry in Equation 10. Finally, the current normal vector $\mathbf{m}$ of the ground plane is shaped by the cross product of points as below:
\begin{flalign}
    \mathbf{m} &= (\mathbf{P'^{}_{i}}-\mathbf{P'^{}_{j}})\times(\mathbf{P'^{}_{i}}-\mathbf{P'^{}_{k}})
\end{flalign}
The ground plane tracing mode is only triggered when the camera is over 300 meters above ground so that the variance of uneven grassland can be approximated by a flat plane compared to the height. 
Next, the rotation matrix to align the normal vector $\mathbf{m}$ in the current frame and the reference normal $\mathbf{n}$ at the start can be derived as follows.
\begin{flalign}
s &= \frac{\mathbf{m}}{||\mathbf{m}||} \cdot \frac{\mathbf{n}}{||\mathbf{n}||}, \\ 
\mathbf{k} &= \frac{\mathbf{m}}{||\mathbf{m}||} \times \frac{\mathbf{n}}{||\mathbf{n}||}, \\
\mathbf{k_{\times}} &= 
\begin{bmatrix}
0 & -k_{3} & k_{2} \\
k_{3} & 0 & -k_{1} \\
-k_{2} & k_{1} & 0
\end{bmatrix}  \\
\mathbf{R} &= \mathbf{I} + \mathbf{k_{\times}} + \mathbf{k_{\times}^{2}} \frac{1}{1+s}  
\end{flalign}
$\bf{k_{\times}}$ is the skew matrix, where non-zero elements are in off-diagonal positions, corresponding to the components of the cross product of $\mathbf{m}$ and $\mathbf{n}$. $s$ is a scale derived from the dot product of two vectors. The rotation matrix is calculated following Rodrigues' rotation formula in Equation 16.

In the end, the 3D Euler angles are retrieved from the rotation matrix according to Equations 17-20, in a right-hand order, "yaw$\leftarrow$pitch$\leftarrow$roll". Only roll and pitch are used for later fusion. 
\begin{flalign} \label{eq:euler}
\mathbf{R} &= 
\begin{bmatrix}
{r}^{}_{11} & {r}^{}_{12} & {r}^{}_{13} \\
{r}^{}_{21} & {r}^{}_{22} & {r}^{}_{23} \\
{r}^{}_{31} & {r}^{}_{32} & {r}^{}_{33}
\end{bmatrix}  \\
\alpha &= \arctan(r^{}_{32}, r^{}_{33})  \\
\beta &= \arctan(-r^{}_{31}, \sqrt{r_{32}^2 + r_{33}^2}) \\
\gamma &= \arctan(r^{}_{21}, r^{}_{11})
\end{flalign}

\section{Adaptive resolution based Particle Filter}
\label{section-pf}
The Particle Filter is easy to implement and applicable in non-linear problems, in particular for positioning \cite{pf-local}. Here, a variant of the vanilla particle filter \cite{pf-tracking}, sampling on the spherical surface adaptively is proposed in the pseudo-code below. In a real configuration, the roll and pitch are virtually constrained by a limit range due to mechanical kinematics, e.g., roll in a range from -45\degree to 45\degree.
\vspace{-1.0em}
\begin{algorithm}[ht]
	\caption{Particle Filter on Spherical Surface}
		 \KwData{${S}_{I}= ({\alpha}_{I}, {\beta}_{I}), {S}_{{C}_{1,2}}= ({\alpha}_{{C}_{1,2}}, {\beta}_{{C}_{1,2}}), \Omega_{1,2,3}$.}
		\KwResult{\textbf{Output} $\bar{S}_{k} = [{\alpha}$, ${\beta}]^T$.}
		\If{new ${S}^{I}$ available}{
        Initialize the particle $s_{0}^{(j)}$ or sampling $s_{k}^{(j)}$ (on $\Omega_{1}$) from $\mathcal{N}(\tilde{\mu}_{I}|\mu_{I} + \omega\delta t,(\sigma_{I}+b))$, $j=1...N$;} 
		\If {new ${S}_{{C}_{1}}$ and ${S}_{{C}_{2}}$ both are available}
		{\For {$s_{k}^{(j)}$ in the particle set of $N$ samples}
		{$\omega_{k}^{(j)}=\omega_{k-1}^{(j)}\exp{(s_{k}^{(j)}-{S}_{{C}_{1,2}})}$; \\ 
		\If {$||s_{k}^{(j)}$-${S}_{{C}_{1,2}}||_{2} \leqslant \epsilon$}
		{Sampling $\hat{s}_{k}^{(1...m)}$ from $\mathcal{N}(\tilde{\mu}_{f}|\mu_{f}, \delta_{f})$ (on $\Omega_{3}$)\\
		initialize new weights: $\omega_{k}^{(1...m)}=\omega_{k}^{(j)}\exp{(\hat{s}_{k}^{(1...m)}-\mu_{f})}$;
		}
		}
		}
		\If {new ${S}_{{C}_{1}}$ or ${S}_{{C}_{2}}$ is available}
		{\For {$s_{k}^{(j)}$ in $N$ samples (on $\Omega_{2}$ resolution)}
		{
		 Sampling from $\mathcal{N}(\tilde{S}_{{C}_{1}}|{S}_{{C}_{1}}, \delta_{{C}_{1}})$ or $\mathcal{N}(\tilde{S}_{{C}_{2}}|{S}_{{C}_{2}}, \delta_{{C}_{2}})$, same as line 6-9; \\
		}
	    }
		$\hat{s}_{k}^{(1...m)} \cup S_{\Omega}$; \\
		Resampling $s_{k}^{(j)}$ according to $\omega_{k}^{(j)}$; \\
		$\bar{S}_{k} = \sum_{j=1}^{N+m}\omega^{(j)}_{k}\hat{s}_{k}^{(j)}$; \\
    \label{alg:pf}
\end{algorithm}
\vspace{-1.0em}
The general idea behind this adaptive particle filter is straightforward. The filter mainly comprises three steps: sampling from orientation measurements of the IMU ${s}_{I}^{}$ (line 2); sampling according to observation from the Computer Vision (CV) pipeline (lines 8 and 15); resampling proportionally to the updated weight of each particle (line 19). The weight in line 6 is derived from a normal distribution, as a function of the square root of the angular distance, which is between sampled cell position and sensor observation on the manifold surface. ${S}_{{C}_{1}}$ and ${S}_{{C}_{2}}$ indicate the orientation estimation from the skyline and the ground plane respectively. 

It is noteworthy that all the particle samples are generated on the discretized cells, spreading over the manifold space formed by the roll and pitch angles. Each particle is a 2D vector represented by a cell position on the spherical surface. There are three levels of cell resolution ($\Omega_{1, 2, 3}$ in Algorithm 1) from coarse to fine, where the longitudinal direction of the spherical surface represents the pitch, while the latitudinal direction is the roll. In line 2, particles are sampled from a Gaussian distribution $\mathcal{N}$, with a mean value at the IMU measurements plus a shift by a constant angular velocity propagating through a certain interval plus an offset $b$. 

When both observations from the skyline and ground plane are available (line 4), more samples will be created around those cells close to sensor measurements (line 8), and their weights are initialized by multiplication of parent weight and local weight as a function of angular distance (line 9), whereas the other particles' weights will be down-weighed by an aforementioned normal distribution as a function of angular distance. Line 7 manifests the angular distance criteria for neighboring particle cells close to observation. The sample cells meeting the criteria are used as parents to create more children particles around $\mu_{f}$ at line 8 of Algorithm 1, and $\mu_{f}$ is derived from Equation \ref{eq:fusion}, as a weighted sum of results from two Computer Vision pipelines, with each result used as a mean value of the normal distribution. A simple inverse of corresponding variance $\delta_{{C}_{1}}, \delta_{{C}_{2}}$ respectively is considered as weight. 
\begin{flalign}\label{eq:fusion}
    \mu_{f} &= \frac{\mu_{{C}_{1}}}{\delta_{{C}_{1}}} + \frac{\mu_{{C}_{2}}}{\delta_{{C}_{2}}} \\
    \delta_{f} &= {\left(\frac{1}{\delta_{{C}_{1}}} + \frac{1}{\delta_{{C}_{2}}}\right)}^{-1}
\end{flalign}
IMU and CV observation variances are set as a constant according to practical tests. $\mu_{f}$ and $\delta_{f}$ are fused results from two CV pipelines in the same weighted sum form of Equations \ref{eq:fusion}. The same strategy repeats when a single CV observation pipeline is present (line 15), but sampling rather on a middle-level resolution.

\begin{figure}[H]
\vspace{-1.2em}
\centering
\includegraphics[width=0.57\textwidth]{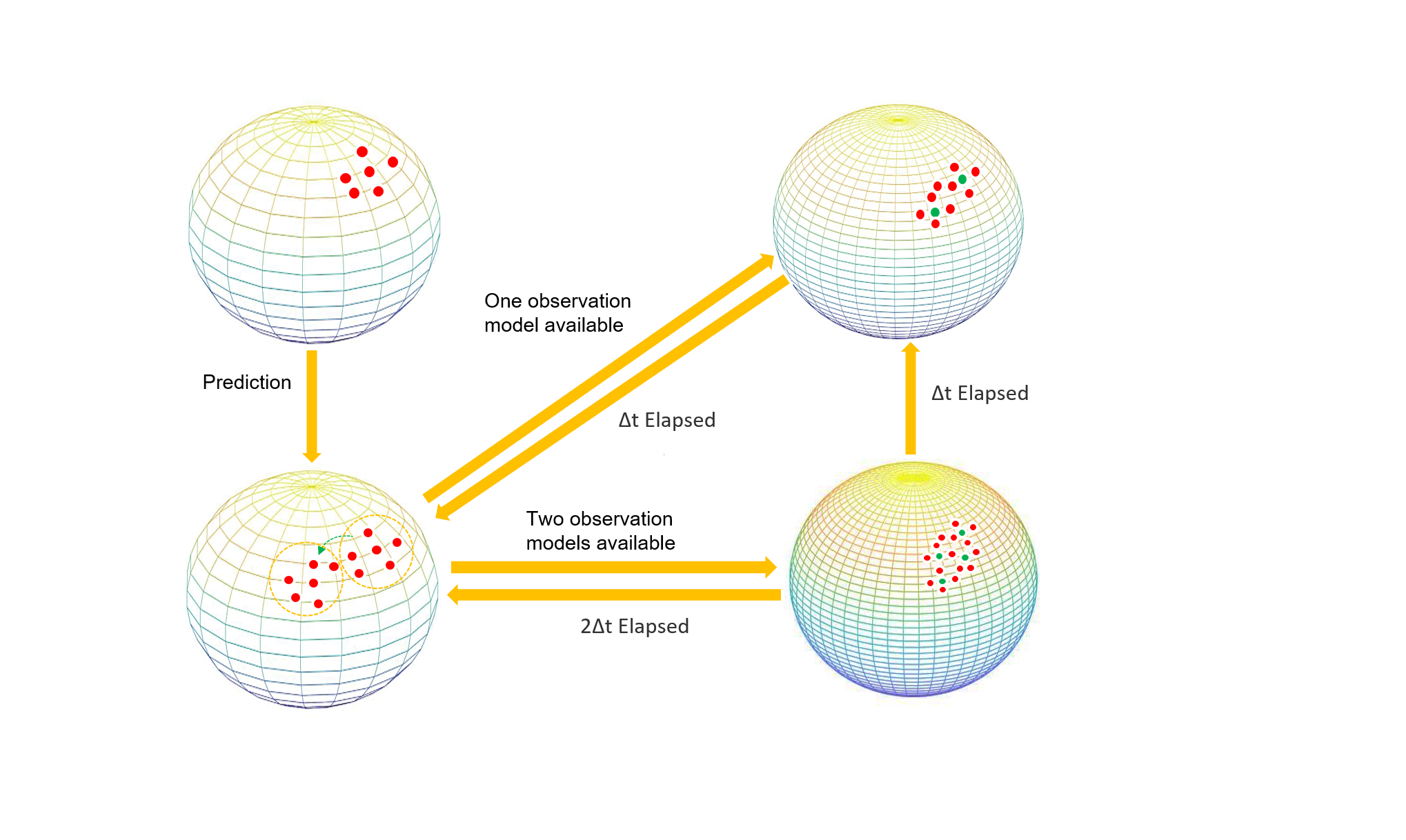}
\vspace{-2.5em}
\caption{Lifetime phases of particle filter sampling on a spherical surface.}
\label{fig:pf-state}
\vspace{-1.3em}
\end{figure}

Each particle's life cycle can be represented in four phases, as shown in Figure \ref{fig:pf-state}. Arrows between them stand for transition conditions. In our setting, the Computer Vision based orientation observation has a smaller variance compared to IMU. As aforementioned, three levels of cell resolution are employed. The top left part of Figure \ref{fig:pf-state} represents the initial state, sampled from a normal distribution centered at the measurement of timestamp $t$. The bottom left is the prediction based on the propagation of the previous particle states by extrapolation in time. When the orientation from a single pipeline, either skyline or ground plane is available, the new samples in red are generated on finer resolution neighboring the green dots corresponding to $\mu_{f}$ in Equation \ref{eq:fusion}. Each dot is located at the center of a cell. If both skyline and ground plane pipelines are available, the highest resolution $\Omega_{3}$ is employed to generate more new samples. Each particle's timestamp of creation is kept as well. If the lapsed time exceeds a certain interval $\delta t$, the particles will be placed back at a coarse resolution, like the arrow direction from the bottom right to the top right. At a certain time point, the particles in the set have various precision. A lifetime check will be called periodically to eliminate the particles existing for a long time.

\section{Experiment \& Results}
\begin{figure}[H]
\vspace{-0.4em}
\centering
\includegraphics[width=0.35\textwidth, scale =0.45]{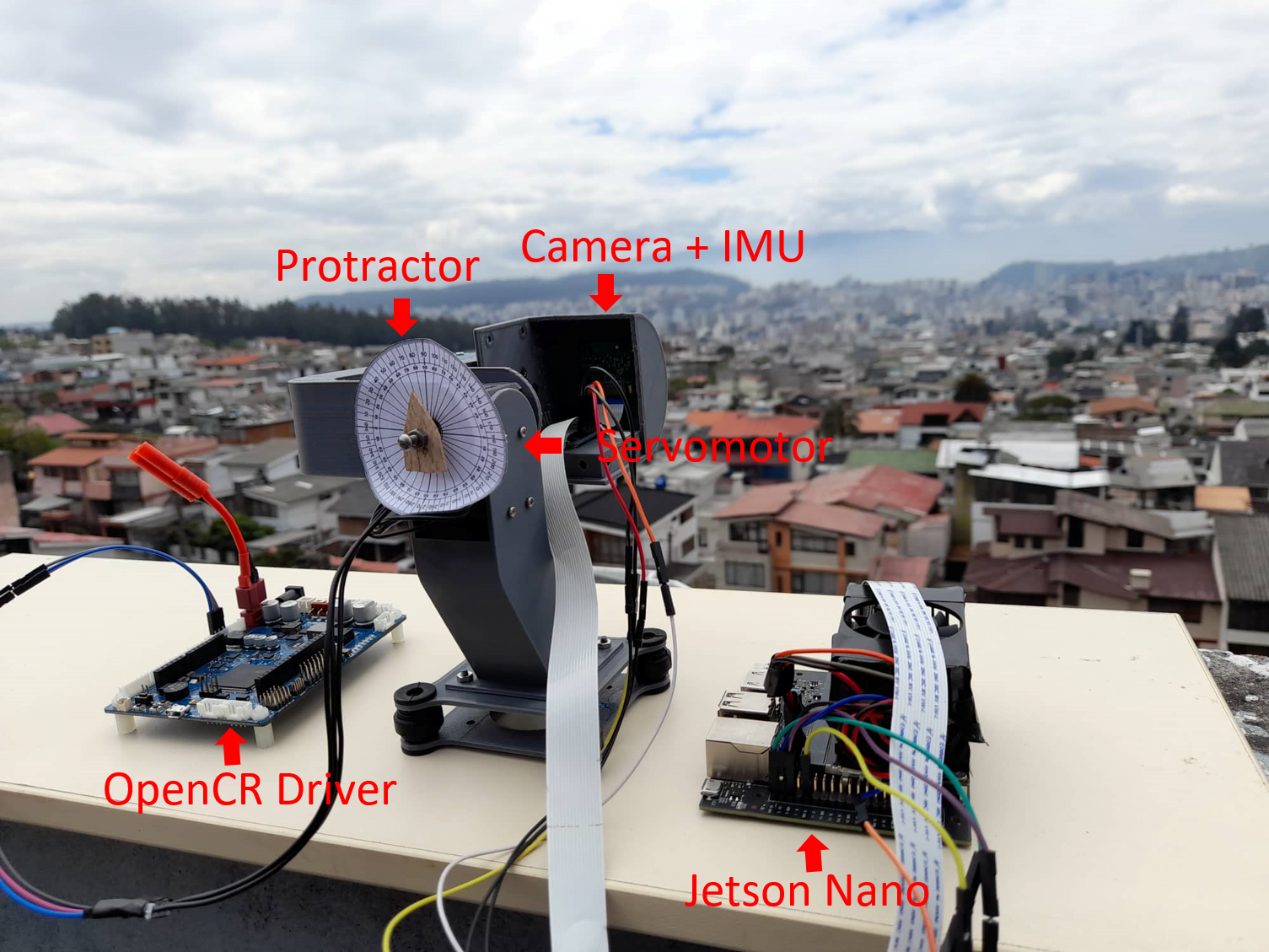}
\caption{Gimbal simulation test setup on top of the building.}
\label{fig:sim-setup}
\vspace{-1.3em}
\end{figure}

In practice, \xk{the video is scaled down to 640$\times$480 resolution to achieve a raw frame rate of 20, while the overall frame rate scales down to 12-15 after the fusion on Jetson Nano}. Intrinsics of the camera are acquired following the calibration guide of \cite{zhang}. Extrinsic calibration between low-cost IMU (BNO055) and Raspi-camera is established using an open source tool "Kalibr" \cite{kalibr1}\cite{kalibr2}. Figure \ref{fig:sim-setup} shows the simulation setup on top of the building in the landscape. There are three parts, a motor driver board, Jetson Nano, and the gimbal part. All 3D-printed cases have enhanced connections, taking aerodynamics into account for flying efficiency. The camera on the gimbal is placed to be forward facing the landscape. Then the gimbal cases along with the sensors are attached to a pole end (not in the view of Figure \ref{fig:sim-setup}). The other end of the pole is controlled manually to simulate random rotation. Here, the ground truth roll and pitch angles are read from the servo motors, and the protractor in the figure is only adopted for verification of the test. In the demo test, we use the fused estimation from our particle filter to steer the motors. Closed-loop PID controller is leveraged for actuation. All the following test sequences were recorded from a static position at the start. Furthermore, it is guaranteed the ground plane should be orthogonal to the gravitational vector at start. This configuration remains unchanged for the real UAV test. 

\xk{Open-source datasets for orientation estimation research were mostly overlapping with SLAM research, and the SLAM datasets were often captured indoors or within urban region, rarely including unpopulated areas, viewed from the top. We thus recorded sequences by using the aforementioned gimbal setup in a real mountain landscape on a tall building roof nearby, which should be quite similar to the camera view on the airplane. During recording, each pose configuration of the gimbal is kept on par with angle readings from motors containing hall sensors as ground truth. For comparisons, the SOTA visual-inertial frameworks "ORBSLAM3" \cite{ORBSLAM3}, "R-VIO" \cite{RVIO}, and "DM-VIO" \cite{DMVIO} were selected at first, but we found these algorithms are dedicated to 6-DoF visual odometry and are all relying on feature points extracted from the images, which are not suitable for the challenging feature-less scenes of our datasets, so we compare our fusion algorithm against the IMU filtered by quaternion-based Madgwick, CV only pipelines, like skyline or ground respectively.}
\begin{table}[H]
\vspace{-0.6em}
\centering
\begin{adjustbox}{width=\columnwidth}
\begin{tabular}{c c c c c c c}
\hline
  &Sequence& IMU (Madgwick Filter) & Skyline only & Ground plane only & Fusion\\
 \hline
\multirow{3}{*}{Roll test (125s)} & test01 &0.0121 & 0.0182 & 0.0344 & \bf{0.0090}\\
& test02 & 0.0147 & 0.0236 & 0.0457 & \bf{0.0126}\\
& test03 & 0.0162 & 0.0325 & 0.0593 & \bf{0.0152}\\
\hline
\hline
\multirow{3}{*}{Pitch test (127s)} & test01 & 0.0147 & 0.0196 & 0.0325 & \bf{0.0118}\\
& test02 & 0.0174 & 0.0214 & 0.0291 & \bf{0.0139}\\
& test03 & 0.0208 & 0.0241 & ----- & \bf{0.0165}\\
\hline
\hline
\multirow{3}{*}{Mixed test (960s)} & test01 & 0.386 & 0.0713 & 0.0674 & \bf{0.0451}\\
& test02 & 0.415 & 0.0651 &  ----- & \bf{0.0584}\\
& test03 & 0.491 & 0.0742 & ----- & \bf{0.0617}\\
\hline
\end{tabular}
\end{adjustbox}
\caption{Average RMSE of roll and pitch angles (in radians). The frame-wise error bigger than radians 0.3 occurring over the half sequence length is considered as failure.}
\label{cmp-result}
\vspace{-1.2em}
\end{table}
\xk{We can conjecture from Table \ref{cmp-result} that our fusion approach consistently outperforms the other baseline approaches without fusion by a considerable margin on all sequences. The sequences cover three movement patterns: sequences with pure roll, pure pitch movement, and random rotation on both axes, and each case is implemented at different angular speeds, ordered from three levels, 3, 9, and 15 degrees per second. The good performance with the lowest RMSE in most tests can be attributed to our good assumptions of the environment, skyline, and ground plane in the wild. The IMU results in the table are from the 6-DoF Madgwick quaternion filter \cite{madgwick}, without the use of a magnetometer. This is because in our case, the mountain region with active volcanoes is affected by the disturbances from the earth's magnetism field change. The filtered IMU results are prone to drift, as presented in the mixed test of 16 minutes, and the errors are nearly one order of magnitude bigger than roll and pitch test sequences. Either the use of skyline or the ground plane as a tracking cue can guarantee the error remaining at a lower level compared to Madgwick filter over IMU measurements, but in some fast rotation cases, the ground planes are partially or not present in the image, which may result in the failure. All of the results in the table justify the merits of using an adaptive particle filter over the manifold, improving the robustness, sensor redundancy, and accuracy.}

\begin{figure}[!thbp]
\vspace{.6em}
\centering
\begin{subfigure}{.48\textwidth}
  \centering
  \includegraphics[width=1.0\textwidth, scale=0.5]{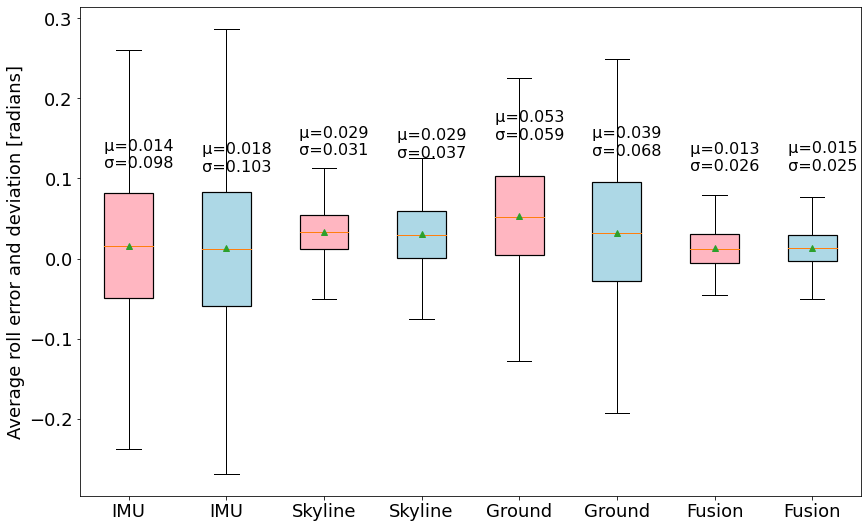}
\end{subfigure}
\caption{Green arrow is mean error, the orange line is a median error, and the box bounds represent the min/max errors. Roll and pitch results are in pink and blue respectively.}
\label{fig:bar-box}
\vspace{-1.0em}
\end{figure}

\xk{Figure \ref{fig:bar-box} further validates the consistency of our method. The test is repeated 10 times per sequence, and then an average error of all trials is taken over the mean error of the whole sequence. The slowest angular speed of the sequence (test01) for pure roll or pitch in Table \ref{cmp-result} is employed. The variances of our fusion results are always the smallest compared to the results of the single sensor modality.}


\section{Conclusion \& Outlook}
A new stand-alone gimbal system is proposed in this paper, based on tracking of natural geometry primitives, skyline, and ground plane approximations. The current frame rotation with respect to a reference frame can be derived by using two lines and normal vectors. Next, a specific particle filter with adaptive resolution-based sampling can fuse orientation from both CV and IMU pipelines, according to the various lifetime phases of one particle. The final experimental results are implemented on a 3D-printed gimbal platform. All simulation tests in the landscape are performed in real-time on a Jetson Nano platform. Single sensor modality-based gimbal solution like IMU is chosen as references, revealing the best accuracy and robustness to drift and disturbances of our approach amongst all comparisons.

Our approach depends on a simple geometric primitive assumption. In challenging weather conditions, abrupt illumination changes, and cloud occlusions pose great challenges to feature tracking-based methods. Our gimbal system cannot work in a different scenario where the skyline is invisible, so for increased robustness, a hybrid system could be applied to general scenes by fusing feature points and the skyline. The estimated skyline is now simplified into a straight line for easy matching and real-time performance, on an affordable edge device. In reality, the curved mountains in the view challenge this assumption. The traditional iterative closest point (ICP) can be applied to the image level for matching. 
In addition,  a Fisheye camera with a full view or multiple cameras at different views should be beneficial to the robustness of the gimbal platform.
\addtolength{\textheight}{-2cm}   

\section{Appendix}
\label{appendix}
Dataset link: \href{https://peridot-sailor-9cd.notion.site/Adaptive-Sampling-based-Particle-Filter-for-Visual-Inertial-Gimbal-in-the-Wild-Datasets-4c8f3a2d27e54ab0871d930571f365ee}{https://peridot-sailor-9cd.notion.site/}
\section*{ACKNOWLEDGMENT}
We thank our colleague Maarten Vandersteegen for the setup advice. We are grateful to the CSC scholarship to fund Xueyang for his research, the EPN international mobility scholarship to fund Ariel for his internship in Belgium, and research Project PIM21-01 as funding for the flight tests.

\newpage

\end{document}